\documentclass[10pt,twocolumn,letterpaper]{article}

\usepackage{cvpr}
\usepackage{times}
\usepackage{epsfig}
\usepackage{graphicx}
\usepackage{amsmath}
\usepackage{amssymb}
\usepackage{mathtools}
\usepackage{multirow}
\usepackage{enumitem}
\usepackage{color}

\newcommand{\keypoint}[1]{\vspace{0.1cm}\noindent\textbf{#1}\quad}
\newcommand{\cut}[1]{}

\makeatother
\usepackage[pagebackref=true,breaklinks=true,letterpaper=true,colorlinks,bookmarks=false]{hyperref}

\cvprfinalcopy 

\def\cvprPaperID{4367} 

\ifcvprfinal\fi
\begin{document}

\title{Sketch Less for More: On-the-Fly Fine-Grained Sketch Based Image Retrieval}
\author{Ayan Kumar Bhunia\textsuperscript{1} \hspace{.2cm} Yongxin Yang\textsuperscript{1} \hspace{.2cm} Timothy M. Hospedales\textsuperscript{1,2}\hspace{.2cm}   Tao Xiang\textsuperscript{1}\hspace{.2cm}  Yi-Zhe Song\textsuperscript{1} \\
\textsuperscript{1}SketchX, CVSSP, University of Surrey, United Kingdom \hspace{.08cm} \textsuperscript{2}University of Edinburgh, United Kingdom.\\
{\tt\small \{a.bhunia, yongxin.yang, t.xiang, y.song\}@surrey.ac.uk, t.hospedales@ed.ac.uk}
}

\maketitle
\ifcvprfinal\thispagestyle{empty}\fi

\begin{abstract}
Fine-grained sketch-based image retrieval (FG-SBIR) addresses the problem of retrieving a particular photo instance given a user's query sketch. Its widespread applicability is however hindered by the fact that drawing a sketch takes time, and most people struggle to draw a complete and faithful sketch. In this paper, we reformulate the conventional FG-SBIR framework to tackle these challenges, with the ultimate goal of retrieving the target photo with the least number of strokes possible. We further propose an on-the-fly design that starts retrieving as soon as the user starts drawing. To accomplish this, we devise a reinforcement learning based cross-modal retrieval framework that directly optimizes rank of the ground-truth photo over a complete sketch drawing episode. Additionally, we introduce a novel reward scheme that circumvents the problems related to irrelevant sketch strokes, and thus provides us with a more consistent rank list during the retrieval. We achieve superior early-retrieval efficiency over state-of-the-art methods and alternative baselines on two publicly available fine-grained sketch retrieval datasets. 
\end{abstract}


\vspace{-0.3cm}
\section{Introduction}

Due to the rapid proliferation of touch-screen devices, the computer vision community has witnessed significant research progress in sketch-related computer vision problems \cite{yu2016sketch, song2017deep, pang2019generalising, dey2019doodle, dutta2019semantically, collomosse2019livesketch}. Among these methods, sketch-based image retrieval (SBIR) \cite{collomosse2019livesketch, dey2019doodle, dutta2019semantically} has received particular attention due to its potential commercial applications. SBIR was initially posed as a category-level retrieval problem. However, it became apparent that the key advantage of sketch over text/tag-based retrieval was conveying \emph{fine-grained} detail \cite{engilberge2018finding} -- leading to a focus on fine-grained SBIR that aims to retrieve a \emph{particular} photo within a gallery. Great progress has been made on FG-SBIR~\cite{yu2016sketch, song2017deep, pang2019generalising}, but two barriers hinder its widespread adoption in practice -- the time taken to draw a complete sketch, and the drawing skill shortage of the user. Firstly, while sketch can convey fine-grained appearance details more easily than text,  drawing a complete sketch is slow compared to clicking a tag or typing a search keyword. Secondly, although state-of-the-art vision systems are good at recognising badly drawn sketches \cite{sangkloy2016sketchy,yu2016sketchAnet}, users who perceive themselves as someone who ``can't sketch'' worry about getting details wrong and receiving inaccurate results.

\begin{figure}[t]
\begin{center}
  \includegraphics[width=\linewidth]{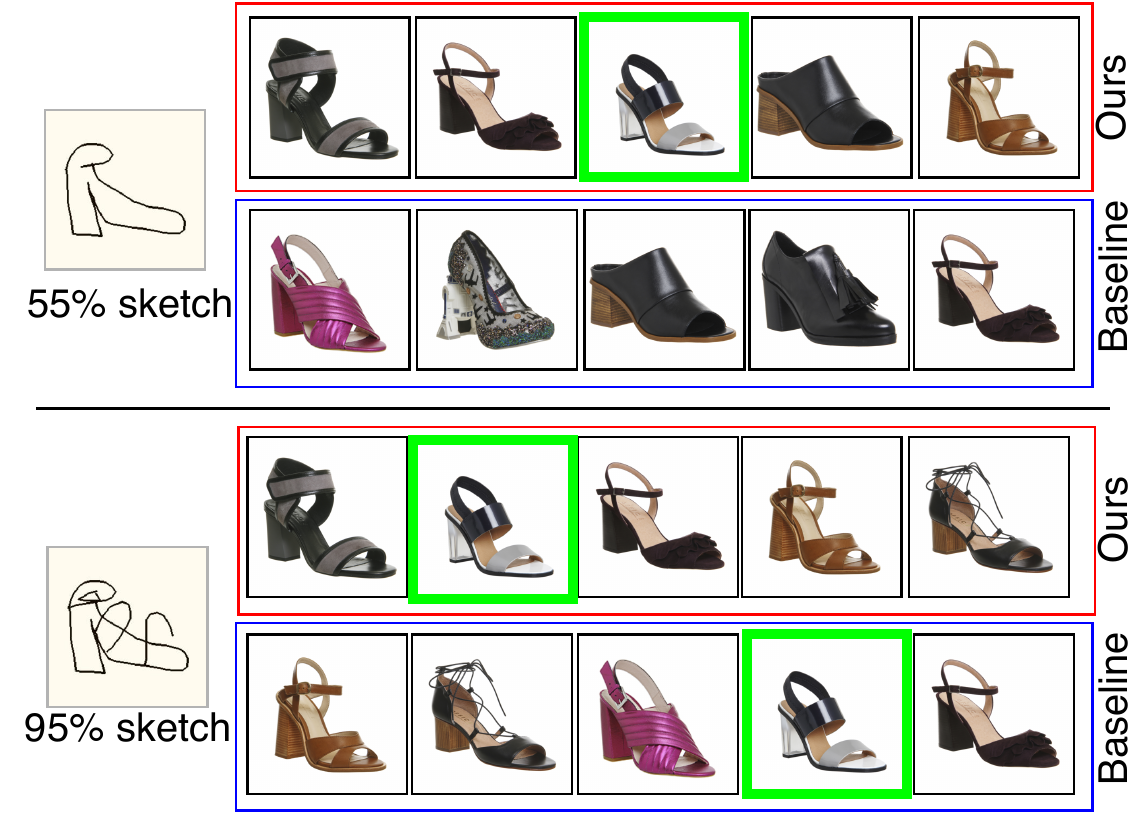}
\end{center}
\vspace{-0.3cm}
  \caption{Examples showing the potential of our framework that can retrieve (top-5 list) target photo using fewer number of strokes than the conventional baseline method.}
  \vspace{-0.45cm}
\label{fig:Fig1_a.pdf}
\end{figure}

In this paper we break these barriers by taking a view of ``less is more'' and propose to tackle a new fine-grained SBIR problem that aims to retrieve the target photo with just a few strokes, as opposed to requiring the complete sketch. This problem assumes a ``on-the-fly'' setting, where retrieval is conducted at every stroke drawn. Figure \ref{fig:Fig1_a.pdf} offers an illustrative example of our on-the-fly FG-SBIR framework. Due to stroke-by-stroke retrieval, and a framework optimised for few-stroke retrieval, users can usually ``stop early'' as soon as their goal is retrieved. This thus  makes sketch more comparable with traditional search methods in terms of time to issue a query, and \emph{more easily} -- as those inexperienced at drawing can retrieve their queried photo based on the easiest/earliest strokes possible \cite{berger2013styleAbstraction}, while requiring fewer of the detailed strokes that are harder to draw correctly.

Solving this new problem is non-trivial.  One might argue that we can directly feed incomplete sketches into the off-the-shelf FG-SBIR frameworks \cite{yu2016sketch, sangkloy2016sketchy}, perhaps also enhanced by including synthesised sketches in the training data. However, those frameworks are not fundamentally designed to handle incomplete sketches. This is particularly the case since most of them employ a triplet ranking framework where each triplet is treated as an independent training example. So they struggle to perform well across a whole range of sketch completion points. Also, the initial sketch strokes could correspond to many possible photos due to its highly abstracted nature, thus more likely to give a noisy gradient. Last, there is no specific mechanism that can guide existing FG-SBIR model to retrieve the photo with minimal sketch strokes, leaving it struggling to perform well across a complete sketching episode during on-the-fly retrieval.

A novel on-the-fly FG-SBIR framework is proposed in this work. First and foremost, instead of the de facto choice of triplet networks that learn an embedding where sketch-photo pairs lie close, we introduce a new model design that optimizes the rank of the corresponding photo over a sketch drawing episode. Secondly, the model is optimised specifically to return the true match within a minimum number of strokes. Lastly, efforts are taken to mitigate the effect of misleading noisy strokes on obtaining a consistent photo ranking list as users add details towards the end of a sketch.

More concretely, we render the sketch at different time instants of drawing, and feed it through a deep embedding network to get a vector representation.  While the other SBIR frameworks \cite{yu2016sketch,sangkloy2016sketchy} use triplet loss \cite{weinberger2009metric_learn_margin} in order to learn an embedding suited for comparing sketch and photo, we optimise the \emph{rank} of the target photo with respect to a sketch query. By calculating the rank of the ground-truth photo at each time-instant $t$ and maximizing the sum of $\frac{1}{rank_t}$ over a complete sketching episode, we ensure that the correct photo is retrieved as early as possible.
Since ranking is a non-differential operation, we use a Reinforcement Learning (RL) \cite{kaelbling1996reinforcement} based pipeline to achieve this goal. Representation learning is performed with knowledge of the whole sequence, as we optimize the reward non-myopically over the sketch drawing episode. This is unlike the triplet loss used for feature learning that does not take into account the temporal nature of the sketch. We further introduce a global reward to guard against harmful noisy strokes especially during later stages of sketching where details are typically added. This also stabilises the RL training process, and produces  smoother retrieval results.

Our contributions can be summarised as follows: (a) We introduce a novel \textit{on-the-fly} FG-SBIR framework trained using reinforcement learning to retrieve photo using an incomplete sketch, and do so with the minimum possible drawing. (b) To this end, we develop a novel reward scheme that models the early retrieval objective, as well as one based on Kendall-Tau \cite{knight1966computer} rank distance that takes into account the completeness of the sketch and associated uncertainty. (c) Extensive experiments on two public datasets demonstrate the superiority of our framework.

\begin{figure*}[t]
	\begin{center}
		\includegraphics[width=1\linewidth]{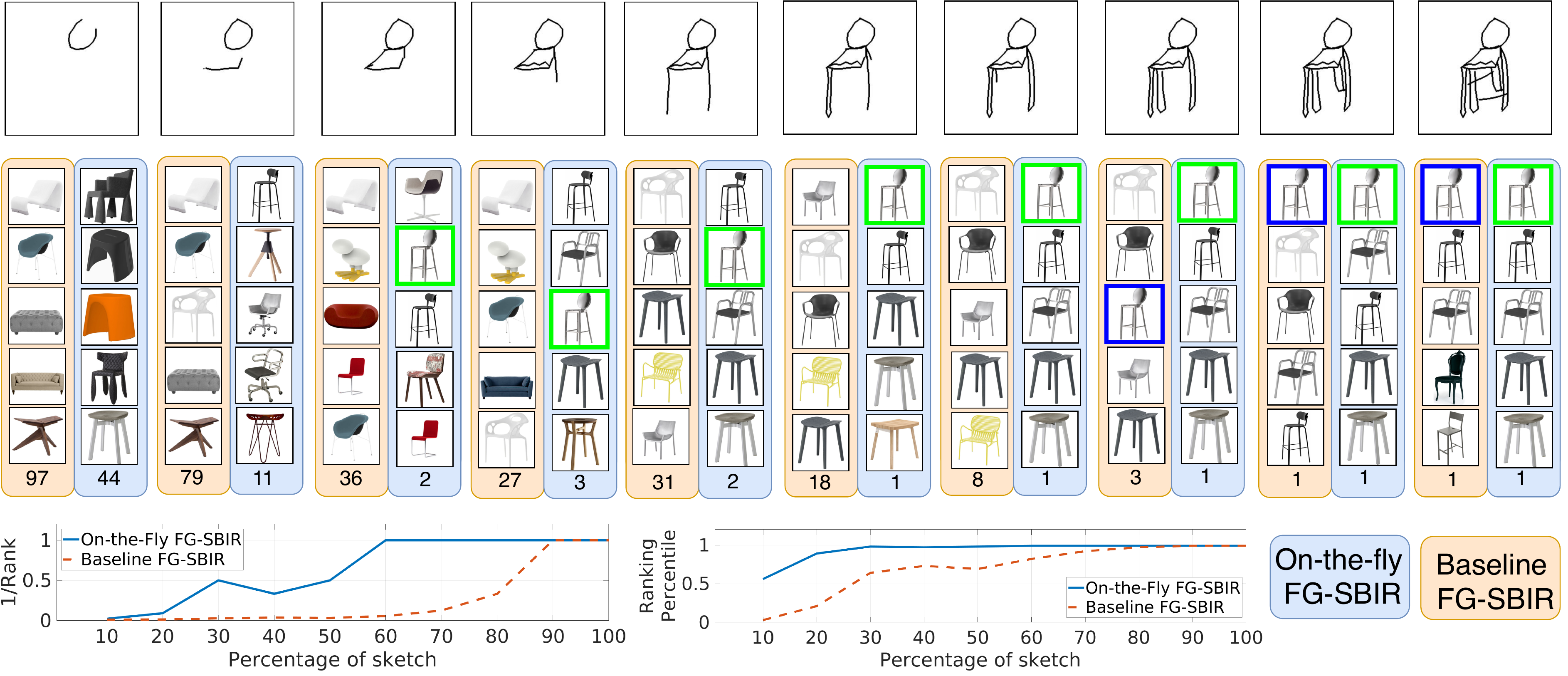}
	\end{center}
	\vspace{-0.25cm}
	\caption{Illustration of proposed \textit{on-the-fly} framework's efficacy over a baseline FG-SBIR method \cite{song2017deep, yu2016sketch} trained with completed sketches only.  For this particular example, our method needs only $30\%$ of the complete sketch to include the true match in the top-10 rank list, compared to $80\%$ for the baseline. Top-5 photo images retrieved by either framework are shown here, in progressive sketch-rendering steps of $10\%$. The number at the bottom denotes the paired (true match) photo's rank at every stage.}
	\label{fig:Fig1}
	\vspace{-0.2cm}
\end{figure*}

\section{Related Works}

\keypoint{Category-level SBIR:} Category-level sketch-photo retrieval is now well studied \cite{cao2011edgel, wang2015sketch, bui2018deep, dey2019doodle, collomosse2017sketching, yelamarthi2018zero, shen2018zero, dutta2019semantically, liu2019semantic, collomosse2019livesketch, liu2017deep}. Contemporary research directions can be broadly classified into traditional SBIR, zero-shot SBIR and sketch-image hashing. In traditional SBIR \cite{cao2011edgel, collomosse2017sketching, collomosse2019livesketch, bui2018deep}, object classes are common to both training and testing. Whereas zero-shot SBIR \cite{yelamarthi2018zero, dey2019doodle, dutta2019semantically, liu2019semantic} asks models to generalise across disjoint training and testing classes in order to alleviate annotation costs. Sketch-image hashing \cite{liu2017deep, shen2018zero} aims to improve the computational cost of retrieval by embedding to binary hash-codes rather than continues vectors.

While these SBIR works assume a single-step retrieval process, a recent study by Collomosse \etal \cite{collomosse2019livesketch} proposed an interactive SBIR framework.
Given an initial sketch query, if the system is unable to retrieve the user's goal in the first try, it resorts to providing some relevant image clusters to the user. The user can now select an image cluster in order to disambiguate the search, based on which the system generates new query sketch for following iteration. This interaction continues until the user's goal is retrieved. This system used Sketch-RNN \cite{ha2017neural} for sketch query generation after every interaction. However Sketch-RNN is acknowledged to be weak in \emph{multi-class} sketch generation  \cite{ha2017neural}. As a result, the generated sketches often diverge from the user's intent leading to poor performance. Note that though such interaction through clusters is reasonable in the case of \emph{category}-level retrieval, it is not applicable to our FG-SBIR task where all photos belong to a single class and differ in subtle ways only.

\keypoint{Fine-grained SBIR:} FG-SBIR is a more recent addition to sketch analysis and also less studied compared to the category-level SBIR task. One of the first studies  \cite{li2014fine} addressed it by graph-matching of deformable-part models. A number of deep learning approaches subsequently emerged \cite{yu2016sketch, song2017deep, pang2017cross, pang2019generalising}. Yu \etal \cite{yu2016sketch} proposed a deep triplet-ranking model for instance-level FG-SBIR. This paradigm was subsequently improved through hybrid generative-discriminative cross-domain image generation \cite{pang2017cross}; and providing an attention mechanism for fine-grained details as well as more sophisticated triplet losses \cite{song2017deep}. Recently Pang \etal \cite{pang2019generalising} studied cross-category FG-SBIR in analogy to the `zero-shot' SBIR mentioned earlier. In this paper, we open up a new research direction by studying FG-SBIR framework design for \emph{on-the-fly and early} photo retrieval.

\keypoint{Partial Sketch:} One of the most popular areas for studying incomplete or partial data is image inpainting \cite{yu2018generative, zheng2019pluralistic}. Significant progress has been made in this area using contextual-attention \cite{yu2018generative} and Conditional Variational Autoencoder (CVAE) \cite{zheng2019pluralistic}. Following this direction, works have attempted to model partial sketch data \cite{liu2019sketchgan, ha2017neural, ghosh2019interactive}. Sketch-RNN \cite{ha2017neural} learns to predict multiple possible endings of incomplete sketches using a Variational Autoencdoer (VAE). While Sketch-RNN works on sequential pen-coordinates, Liu \etal \cite{liu2019sketchgan} extend conditional image-to-image translation to rasterized sparse sketch domain for partial sketch completion, followed by an auxiliary sketch recognition task. 
Ghosh \etal \cite{ghosh2019interactive} proposed an interactive sketch-to-image translation method, which completes an incomplete object outline, and thereafter it generates a final synthesised image. 
Overall, these works first try to complete the partial sketch by modelling a conditional distribution based on image-to-image translation, and subsequently focus on specific task objective, be it sketch recognition or sketch-to-image generation. Unlike these two-stage inference frameworks, we focus on instance-level photo retrieval with a minimum number of sketch strokes, thus enabling partial sketch queries in a single step.

\keypoint{Reinforcement Learning in Vision:} There has been significant progress in leveraging Reinforcement Learning (RL) \cite{kaelbling1996reinforcement} techniques in various computer vision problems \cite{wang2019reinforced, han2019deep}. Vision applications benefiting from RL include visual relationship detection \cite{liang2017deep}, automatic face aging \cite{duong2019automatic}, vision-language navigation \cite{wang2019reinforced} and 3D scene completion \cite{han2019deep}. 
In terms of sketch analysis, RL was leveraged to study abstraction and summarisation by trading off between recognisability of a sketch and number of strokes \cite{riaz2018learning,muhammad2019goal}. While these studies aimed to discover salient strokes by using RL to filter out unnecessary strokes from a given complete sketch, we focus on leveraging RL to retrieve a photo on-the-fly with a minimum number of strokes.  

\begin{figure*}[t]
\begin{center}
  \includegraphics[width=\linewidth]{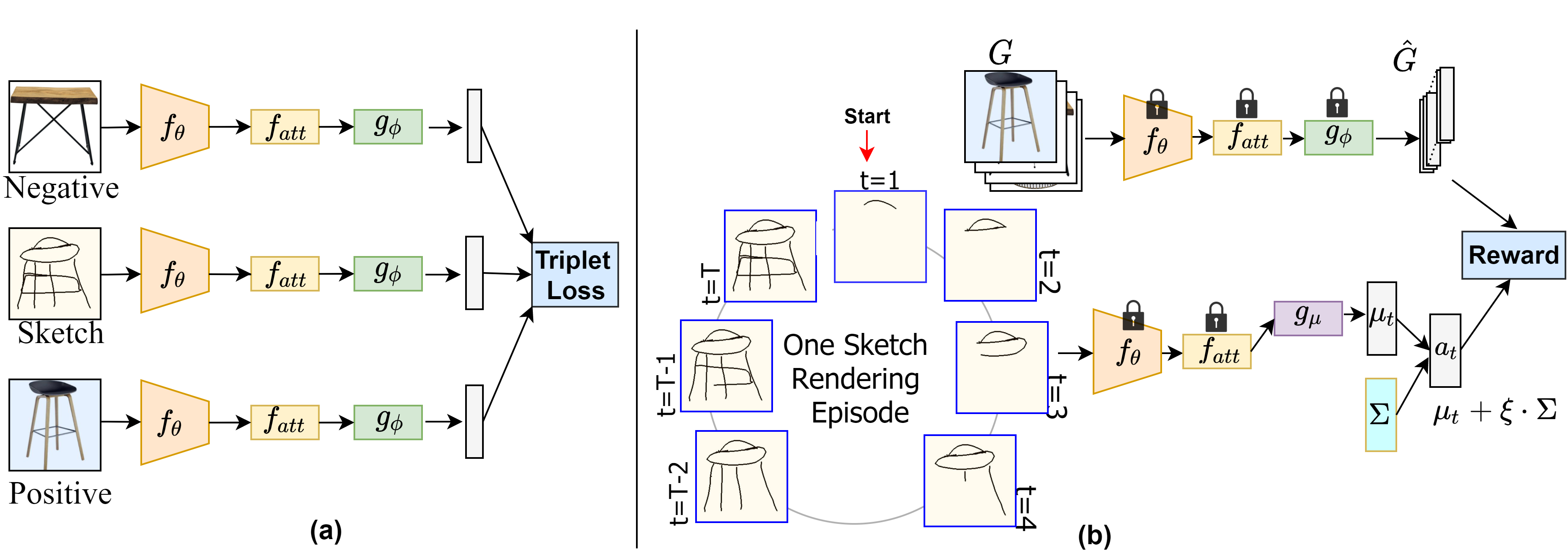}
\end{center}
\vspace{-.15in}
  \caption{(a) A conventional FG-SBIR framework trained using triplet loss. (b) Our proposed reinforcement learning based framework that takes into account a complete sketch rendering episode. Key locks signifies particular weights are fixed during RL training.}
\vspace{-.18in}
\label{fig:Fig2}
\end{figure*}

\section{Methodology}

\keypoint{Overview:}
Our objective is to design an `on-the-fly' FG-SBIR framework, where we perform live analysis of the sketch as the user draws. The system should re-rank candidate photos based on the sketch information up to that instant and retrieve the target photo at the earliest stroke possible (see Figure \ref{fig:Fig1} for an example of how the framework works in practice). To this end, we first pre-train a state-of-the-art FG-SBIR model \cite{yu2016sketch, song2017deep} using triplet loss. Thereafter, we keep the photo branch fixed, and fine-tune the sketch branch through a non-differentiable ranking based metric over complete sketch drawing episodes using reinforcement-learning.

Formally, a pre-trained  FG-SBIR model learns an embedding function $F(\cdot): I \rightarrow \mathbb{R}^{D}$ mapping a rasterized sketch or photo $I$ to a $D$ dimensional feature. Given a gallery of $M$ photo images $G = \left \{ X_{i} \right \}_{i=1}^{M}$, we obtain a list of $D$ dimensional vectors $\hat{G}= \left \{ F(X_{i}) \right \}_{i=1}^{M}$ using $F(.)$. Now, for a given query sketch $S$, and some pairwise distance metric, we obtain the top-$q$ retrieved photos from $G$, denoted as $Ret_{q}(F(S), \hat{G})$. If the ground truth (paired) target photo appears in the top-$q$ list, we consider top-$q$ accuracy to be true for that sketch sample. Since we are dealing with on-the-fly retrieval, a sketch is represented as $S \in \left \{p^{1}, p^{2}, p^{3}, ... p^{N}  \right \}$, where $p^{i}$ denotes one sketch coordinate tuple $(x,y)$, and $N$ stands for maximum number of points. We assume that there exists a sketch rendering operation $ \varnothing $, which takes a list $S^{K}$ of the \textit{first $K$ coordinates} in $S$, and produces one rasterized sketch image. Our objective is to train the framework so that the ground-truth paired photo appears in $Ret_{q}(F(\varnothing(S^{K})), \hat{G})$ with a minimum value of $K$.

\subsection{Background: Base Models} \label{basemodel}
For pre-training, we use a state-of-the-art Siamese network \cite{song2017deep} with three CNN branches with shared weights, corresponding to a query sketch, positive and negative photo respectively (see Figure \ref{fig:Fig2}(a)). Following recent state-of-the-art sketch feature extraction pipelines \cite{dey2019doodle, song2017deep}, we use soft spatial attention \cite{xu2015show} to focus on salient parts of the feature map. Our baseline model consists of three specific modules: (a) $f_{\theta}$ is initialised from pre-trained InceptionV3 \cite{szegedy2016rethinking} weights, (b) $f_{att}$ is modelled using 1x1 convolution followed by a softmax operation, (c) $g_{\phi}$ is a final fully-connected layer with $l_2$ normalisation to obtain an embedding of size D. Given a feature map $B = f_{\theta}(I)$, the output of the attention module is computed by $B_{att} = B + B\cdot f_{att}(B) $. Global average pooling is then used to get a vector representation, which is again fed into $g_{\phi}$ to get the final feature representation used for distance calculation. We considered $f_{\theta}$, $f_{att}$, and $g_{\phi}$ to be wrapped as an overall embedding function $F$. The training data are triplets $\{a, p, n\}$ containing sketch anchor, positive and negative photos respectively. The model is trained using \textit{triplet loss} \cite{weinberger2009metric_learn_margin} that aims to reduce the distance between sketch anchor and positive photo $\beta^{+} = \left \| F(a) -F(p) \right \|_{2}$, while increasing the distance between sketch anchor and negative photo $\beta^{-} = \left \| F(a) -F(n) \right \|_{2}$. Hence, the triplet loss can be formulated as $max\{0, \mu + \beta^{+} - \beta^{-}\}$, where $\mu$ is the margin hyperparameter.

\subsection{On-The-Fly FG-SBIR}

\keypoint{Overview:} We model an on-the-fly FG-SBIR model as a sequential decision making process \cite{kaelbling1996reinforcement}. Our agent takes \emph{actions} by producing a feature vector representation of the sketch at each rendering step, and is \emph{rewarded} by retrieving the paired photo early. Due to computation overhead, instead of rendering a new sketch at every coordinate instant, we rasterize the sketch a total of $T$ times, i.e., at steps of interval $\left \lfloor \frac{N}{T} \right \rfloor$. As the photo branch remains constant, we get $\hat{G}$ using the baseline model. We train the agent (sketch-branch) to deal with partial sketches. In this stage we fine-tune the sketch branch only, aiming to make it competent in dealing with partial sketches. 
Considering one sketch rendering episode as $S \in \{ p_{1}, p_{2}, p_{3}, ..., p_{T} \}$, the agent takes state $s_{t} = \varnothing(p_{t})$ as input at every time step $t$, producing a continuous `action' vector $a_{t}$. Based on that, the retrieval environment returns one reward $r_{t}$, mainly taking into account the pairwise distance between $a_{t}$ and $\hat{G}$. The goal of our RL model, is to find the optimal policy for the agent that maximises the total reward under a complete sketch rendering episode.

Triplet loss \cite{song2017deep,weinberger2009metric_learn_margin} considers only a single instant of a sketch. However, due to creation of multiple partially-completed instances of the same sketch, a diversity is created which confuses the triplet network. In contrast, our approach takes into account the complete episode of progressive sketch generation before updating the weights, thus providing a more principled and practically reliable way to model partial sketches.



\keypoint{Model:} The sketch branch acts as our agent in RL framework, based on a stochastic continuous Gaussian policy \cite{duan2016benchmarking}, where action generation is represented with a multivariate Normal distribution. Following the typical RL notation, we define our policy as $\pi_{\Theta }(a|s)$. $\Theta$ encases the parameters of policy network comprised of pre-trained $f_{\theta}$ and $f_{att}$ which remains fixed, and a fully-connected trainable layer $g_{\mu}$ that finally predicts the mean vector \textbf{$\mu$} of the multivariate Gaussian distribution. Please refer to Figure \ref{fig:Fig2}(b) for an illustration. At each time step $t$, a policy distribution $\pi_{\Theta }(a_{t}|s_{t})$ is constructed from the  distribution parameters predicted by our deep network. Following this, an action is sampled from this distribution, acting as a $D$ dimensional feature representation of the sketch at that instant, i.e. $a_{t} \sim \pi_{\Theta }(\cdot |s_{t})$. Mathematically, this Gaussian policy is defined as:
\vspace{-0.3cm}
 \begin{multline}\label{policy_eqn}
 \mathrm{\pi_{\Theta }(a_{t}|s_{t}) = \sqrt{\frac{1}{(2\pi)^{D} \left | \Sigma \right |}} \times} \\  \mathrm{exp\left \{-\frac{1}{2}(a_{t} - \mu_{t})^{\top} \Sigma^{-1} (a_{t} - \mu_{t}) \right \}}
 \end{multline}

\noindent where the mean ${\mu_{t}} = g_{\mu}(s'_{t}) \in \mathbb{R}^{D}$, and $s'_{t}$ is obtained via a pre-trained  $f_{\theta}$ and $f_{att}$ that take state $s_{t} = \varnothing(p_{t})$ as its input. Meanwhile, $\Sigma$ is a standalone trainable diagonal covariance matrix. We sample action $a_{t} = \mu_{t} +  \xi \cdot \Sigma $, where $\xi \sim \mathcal{N}(\mathbf{0},\mathbf{I})$ and $a_{t} \in \mathbb{R}^{D}.$

\keypoint{Local Reward:} In line with existing works leveraging RL to optimize non-differentiable task metrics in computer vision (e.g.,~\cite{rennie2017self}), our  optimisation objective is the non-differentiable \textit{ranking} metric. The distance between a query sketch embedding and the paired photo should be lower than the distance between the query and all other photos in $G$. In other words, our objective is to minimise the rank of paired photo in the obtained rank list. Following the notion of maximising the reward over time, we maximise the \textit{inverse rank}. For $T$ sketch rendering steps under a complete episode of each sketch sample, we obtain a total of $T$ scalar rewards that we intend to maximize:
\begin{equation}\label{reward_1}
R_{t}^{Local}  =  \frac{1}{rank_{t}}
\end{equation}
From a geometric perspective, assuming a high value of T, this reward design can be visualised as maximising the area under a curve, where the $x$ and $y$ axes correspond to \textit{percentage of sketch} and $\frac{1}{rank_t}$ respectively. Maximising this area therefore requires the model to achieve early retrieval of the required photo.

\keypoint{Global Reward:}  
During the initial steps of sketch rendering, the uncertainty associated with the sketch representation is high because an incomplete sketch could correspond to various photos (e.g., object outline with no details yet). The more it progresses towards completion, the representation becomes more concrete and moves towards one-to-one mapping with a corresponding photo. 
To model this observation, we use Kendall-Tau-distance \cite{knight1966computer} to measure the distance between two rank lists obtained from sequential sketching steps $L_t$ and $L_{t+1}$. Kendall-Tau measures the distance between two ranking lists \cite{pedronette2013image} as the number of pairwise disagreements (pairwise ranking order change) between them.
Given the expectation of more randomness associated with early ambiguous partial sketches, the Kendall-Tau-distance between two successive rank lists from the initial steps of an episode is expected to be higher.
Towards its completion, this value should decrease as the sketch becomes more unambiguous. 
With this intuition, we add a regularizer that encourages the \emph{normalised} Kendall-Tau-distance $\tau$ between two successive rank lists to be monotonically decreasing over a sketch rendering episode: 
\begin{equation}\label{reward_2}
R_{t}^{Global}= -max(0,  \tau(L_{t}, L_{t+1}) -\tau(L_{t-1}, L_{t}))
\end{equation}

This global regularisation reward term serves three purposes: (a) It models the extent of uncertainty associated with the partial sketch. (b) It discourages excessive change in the rank list later in an episode, making the retrieved result more consistent. This is important for user experience: if the returned top-ranked photos are changing constantly and drastically as the user adds more strokes, the user may be dissuaded from continuing.   (c) Instead of simply considering the rank of the target, it considers the behaviour of the full ranking list and its consistency at each rendering step.


\vspace{0.1cm}
\noindent \textbf{Training Procedure:} We aim at maximising the sum of two proposed rewards
\begin{equation}\label{reward_3}
R_{t}  = \gamma_{1} \cdot R_{t}^{Local} + \gamma_{2} \cdot R_{t}^{Global}
\end{equation}

The RL literature provides several options for optimisation. The vanilla policy gradient \cite{mnih2016asynchronous} is the simplest, but suffers from poor data efficiency and robustness. Recent alternatives such as trust region policy optimization (TRPO) \cite{schulman2015trust} are more data efficient, but involves complex second order derivative matrices. We therefore employ Proximal Policy Optimization (PPO) \cite{schulman2017proximal}. Using only first order optimization, it is data efficient as well as simple to implement and tune. PPO tries to limit how far the policy can change in each iteration, so as to reduce the likelihood of taking wrong decisions. More specifically, in vanilla policy gradient, the current policy is used to compute the policy gradient whose objective function is given as:
\begin{equation}\label{ppo_1}
J(\Theta)= \mathrm{\hat{\mathop{\mathbb{E}}}_{t}\left [  log\pi_{\Theta}(a_{t}|s_{t}) R_{t}\right ]}
\end{equation}

\noindent PPO uses the idea of importance sampling \cite{neal2001annealed} and maintains two policy networks, where evaluation of the current policy $\pi_{\Theta}(a_{t}|s_{t})$ is done by collecting samples from the older policy $\pi_{old}(a_{t}|s_{t})$, thus helping in sampling efficiency. Hence, along-with importance sampling, the overall objective function  written as:

\begin{equation}\label{ppo_2}
J(\Theta)= \mathrm{\hat{\mathop{\mathbb{E}}}_{t}\left [  \frac{\pi_{\Theta}(a_{t}|s_{t})}{\pi_{old}(a_{t}|s_{t})} r_{t}\right ]} = \mathrm{\hat{\mathop{\mathbb{E}}}_{t}\left [ m_{t}(\Theta) R_{t}\right ]}
\end{equation}

\noindent where $m_{t}(\Theta)$ is the probability ratio $m_{t}(\Theta) = \frac{\pi_{\Theta}(a_{t}|s_{t})}{\pi_{old}(a_{t}|s_{t})}$, which measures the difference between two polices. Maximising Eq.~\ref{ppo_2} would lead to a large policy update, hence it penalises policies moving $m_{t}(\Theta)$ away from 1, and the new clipped surrogate objective function becomes: 
\begin{equation}\label{ppo_3}
J(\Theta)= \mathrm{\hat{\mathop{\mathbb{E}}}_{t}\left [min(m_{t}(\Theta)r_{t}, clip(m_{t}(\Theta), 1 - \varepsilon, 1 + \varepsilon)R_{t}) \right ]}
\end{equation}
\noindent where $\varepsilon$ is a hyperparameter  set to 0.2 in this work. Please refer to \cite{schulman2017proximal} for more details. Empirically, we found the actor-only version of PPO with clipped surrogate objective to work well for our task. More analysis is given in Sec.~\ref{abla}. 

\section{Experiments}\label{sec:experiments}
\vspace{-0.1cm}
\keypoint{Datasets:} We use QMUL-Shoe-V2 \cite{pang2019generalising, riaz2018learning, song2018learning} and QMUL-Chair-V2 \cite{song2018learning} datasets that have been specifically designed for FG-SBIR.
Both datasets contain coordinate-stroke information, enabling us to render the rasterized sketch images at intervals, for training our RL framework and evaluating its retrieval performance over different stages of a complete sketch drawing episode. QMUL-Shoe-V2 contains a total of 6,730 sketches and 2,000 photos, of which we use 6,051 and 1,800 respectively for training, and the rest for testing. For QMUL-Chair-V2, we split it as 1,275/725 sketches and 300/100 photos for training/testing respectively.

\keypoint{Implementation Details:} We implemented our framework in PyTorch \cite{paszke2017automatic} conducting experiments on a 11 GB Nvidia RTX 2080-Ti GPU. An Inception-V3 \cite{szegedy2016rethinking} (excluding the auxiliary branch) network pre-trained on ImageNet datasets \cite{russakovsky2015imagenet}, is used as the backbone network for both sketch and photo branches. In all experiments, we use Adam optimizer \cite{kingma2014adam} and set $D = 64$ as the dimension of final feature embedding layer. We train the base model with a triplet objective having a margin of $0.3$, for $100$ epochs with batch size $16$ and a learning rate of $0.0001$. During RL based fine-tuning of sketch branch, we train the final $g_{\mu}$ layer of sketch branch (keeping $f_{\theta}$ and $f_{att}$ fixed) with $\Sigma$ for $2000$ epochs with an initial learning rate of 0.001 till epoch 100, thereafter reducing it to 0.0001. The rasterized sketch images are rendered at $T = 20$ steps, and the gradients are updated  by averaging over a complete sketch rendering episode of $16$ different sketch samples. In addition to normalising the sampled action vector $a_{t}$, $l_{2}$ normalisation is also used after global adaptive average pooling layer as well as after the final feature embedding layer $g_{\phi}$ in image branch. The diagonal elements of $\Sigma$ are initialised with $1$, and $\gamma_{1} $,  $\gamma_{2}$  and $\varepsilon$ are set to 1, $\mathrm{1e-4}$  and 0.2 respectively.

\keypoint{Evaluation Metric:} In line with on-the-fly FG-SBIR setting, we consider results appearing at the top of the list to matter more. Thus, we quantify the performance using Acc.@$q$ accuracy, i.e. percentage of sketches having true-match photos appearing in the top-$q$ list. Moreover, in order to capture the early retrieval performance, shadowing some earlier image retrieval works \cite{kovashka2012whittlesearch}, we use plots of (i) \textit{ranking percentile} and (ii) \textit{$\frac{1}{rank}$} versus \textit{percentage of sketch}. In this context, a higher value of the mean area under the curve for (i) and (ii) signifies better early sketch retrieval performance, and we use m@A and m@B as shorthand notation for them in the rest of the paper, respectively.   

\subsection{Baseline Methods}
To the best of our knowledge, there has been no prior work dealing with early retrieval in SBIR. Thus, based on some earlier works, we chose existing FG-SBIR baselines and their adaptations towards the new task to verify the contribution of our proposed RL based solution.

\textbullet \ \textbf{B1}: Here, we use the baseline model \cite{song2017deep, yu2016sketch} trained only with triplet loss. This basically represents our model (see Section~\ref{basemodel}) before RL based fine-tuning.\ \textbullet \  \textbf{B2}: We train a standard triplet model, but use all intermediate sketches as training data, so that the model also learns to retrieve incomplete sketches.\ \textbullet \   \textbf{B3}: We train $20$ different models (as $T=20$) for the sketch branch, and each model is trained to deal with a specific percentage of sketch (like 5\%, 10\%, ..., 100\%), thus increasing the number of trainable parameters $20$ times than the usual baseline. Different models are deployed at different stages of completion -- again not required by any other compared methods.\ \textbullet \  \textbf{B4}: While RL is one of the possible ways of dealing with non-differentiable metrics, the recent work of Engilberge \etal \cite{engilberge2019sodeep} introduced a generalized deep network based solution to approximate non-differentiable objective functions such as ranking. This can be utilized in a plug-and-play manner within existing deep architectures. We follow a similar setup of cross-modal retrieval as designed by Engilberge \etal \cite{engilberge2019sodeep} and impose combination of triplet loss and ranking loss at T different instants of the sketch.

\begin{figure*}[]
	\begin{center}
		\includegraphics[width=1\linewidth]{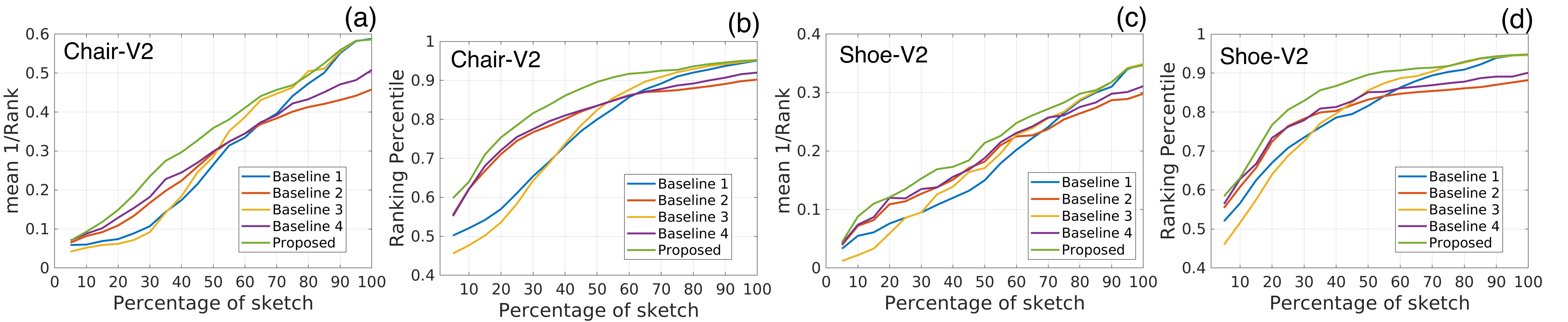}
	\end{center}
	\vspace{-0.15in}
	\caption{Comparative results. Note that instead of showing T=20 sketch rendering steps, we visualize it through percentage of sketch here. A higher area under these plots indicates better \emph{early retrieval performance}.}
	\label{fig:Graph1_Full}
	\vspace{-.15in}
\end{figure*}

\subsection{Performance Analysis}\label{sec:perf}

The performance of our proposed on-the-fly sketch based image retrieval is shown in Figure~\ref{fig:Graph1_Full} against the baselines methods. We observe: 
(i) State-of-the-art triplet loss based baseline \textbf{B1} performs poorly for early incomplete sketches, due to the absence of any mechanism for learning incomplete retrieval. 
(ii) \textbf{B2}'s imposition of triplet-loss at every sketch-rendering step provides improved retrieval performance over \textbf{B1} for a few initial instants, but its performance declines towards the completion of sketch. This is mainly due to the fact that imposing triplet loss over incomplete sketches destabilises the learning process as it generates noisy gradients.
In contrast, our RL based pipeline takes into account a complete sketch rendering episode along-with the associated uncertainty of early incomplete sketches before updating the gradients. 
(iii) Designing 20 different sketch models in \textbf{B3} for $T=20$ different sketch rendering steps, improves performance towards the end of sketch rendering episode after 40\% of sketch rendering in comparison to \textbf{B1}. However, it is poor for sketches before that stage due to its incompleteness which could correspond to various possible photos. 
(iv) An alternative of RL method -- differential sorter described in \textbf{B4}, fares well against baseline \textbf{B1}, but is much weaker in comparison to our RL based non-differentiable ranking method. A qualitative result can be seen in Figure \ref{fig:Fig1} where \textbf{B1} is the baseline.

In addition to the four baselines,  following the recent direction of dealing with partial sketches  \cite{ghosh2019interactive, liu2019sketchgan}, we tried a two stage framework for our early retrieval objective referred as \textbf{TS} in Table \ref{tab:my-table1}. At any given drawing step, a conditional image-to-image translation model \cite{isola2017image} is used to generate the complete sketch. Thereafter, it is fed to an off-the-shelf baseline model for photo retrieval. However, this choice of using an image translation network to complete the sketch from early instances, fails miserably. Moreover, it merely produces the input sketch with a few new random noisy strokes.

To summarise, our RL framework outperforms a number of baselines by a significant margin in context of early image retrieval performance as seen from the quantitative results in Figure~\ref{fig:Graph1_Full} and Table \ref{tab:my-table1}, without deteriorating top-5 and top-10 accuracies in retrieval performance.

\subsection{Ablation Study}\label{abla}
\keypoint{Different RL Methods:} We compare Proximal Policy Optimization (PPO), used here for continuous-action reinforcement learning, with some alternative RL algorithms. Although we intended to use a complete actor-critic version of PPO combining the policy surrogate Eq.~\ref{ppo_3} and value function error \cite{schulman2017proximal} term, using the actor-only version works better in our case. Additionally we evaluate this performance with (i) vanilla policy gradient \cite{mnih2016asynchronous}\cut{ which apparently suffers from poor data efficiency,} and (ii) TRPO \cite{schulman2015trust}\cut{which requires evaluating a complicated second order derivative matrix}. Empirically we observe a higher performance with clipped surrogate objective when compared with the adaptive KL penalty (with adaptive co-efficient 0.01) approach of PPO.  Table~\ref{tab:my-table2} shows that our actor-only PPO with clipped surrogate objective outperforms other alternatives.

\keypoint{Reward Analysis:} In contrast to the complex design of efficient optimization approaches for non-differentiable rank based loss functions \cite{mohapatra2018efficient, engilberge2019sodeep}, we introduce a simple reinforcement learning based pipeline that can optimise a CNN to achieve any non-differentiable metric in a cross modal retrieval task. 
To justify our contribution and understand the variance in retrieval performance with different plausible reward designs, we conduct a thorough ablative study. 
The positive scalar reward value is assigned to $1$ (otherwise zero), when the paired photo appears in top-$q$ list. This $q$ value could be controlled based on the requirements. 
Instead of reciprocating the rank value, taking its negative is also a choice. To address the concern that our inverse rank could produce too small a number, we alternatively evaluate the square root of  reciprocal rank. From the results in Table~\ref{tab:my-table3}, we can see that our designed reward function (Eq.~\ref{reward_3}) achieves the best performance.

 \begin{figure}[t]
\begin{center}
  \includegraphics[width=\linewidth]{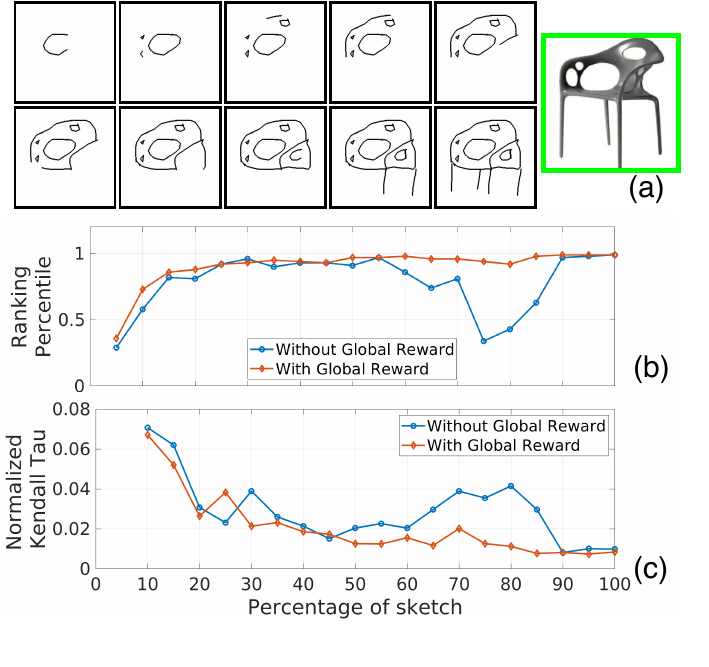}
\end{center}
\vspace{-.30in}
  \caption{
  (a) Example showing progressive order of completing a sketch.
  (b) shows the drop in percentile whenever an irrelevant stroke is introduced while drawing (blue).
  (c) shows the corresponding explosive increase of Kendall-Tau distance signifying the percentile drop (blue). Our global reward term (red) nullifies these negative impacts of irrelevant sketch strokes thus maintaining consistency of the rank-list overall.
  }
 \vspace{-.22in}
\label{fig:Fig3}
\end{figure}

\begin{table}[]
 
\centering
\caption{Comparative results with different baseline methods. Here A@5 and A@10 denotes top-5 and top-10 retrieval accuracy for complete sketch (at t=T), respectively, whereas, m@A and m@B quantify the retrieval performance over a sketch rendering episode (see Section~\ref{sec:experiments} for metric definition).}
 
\label{tab:my-table1}
\resizebox{1.\columnwidth}{!}{
\begin{tabular}{ccccccccc}
\hline
 & \multicolumn{4}{c}{Chair-V2} & \multicolumn{4}{c}{Shoe-V2} \\ \cline{2-9} 
 & m@A & m@B & A@5 & A@10 & m@A & m@B & A@5 & A@10 \\ \hline
B1 & 77.18 &  29.04 & \textbf{76.47}  & 88.13 & 80.12 & 18.05  & 65.69 & \textbf{79.69}  \\ 
B2 & 80.46 &  28.07 &  74.31 &  86.69 &79.72  & 18.75 &  61.79 & 76.64  \\
B3 & 76.99 &  30.27 &  \textbf{76.47} & 88.13 & 80.13 &  18.46 &  65.69 & 79.69   \\ 
B4 & 81.24 &  29.85 &  75.14 &  87.69 & 81.02 & 19.50 & 62.34 &  77.24 \\  
TS & 76.01 &  27.64 & 73.47  & 85.13 & 77.12 & 17.13  & 62.67 & 76.47  \\ 
Ours & \textbf{85.44} &  \textbf{35.09} & 76.34 & \textbf{89.65} & \textbf{85.38} & \textbf{21.44} &   \textbf{65.77}&  79.63  \\ \hline
\end{tabular}%
}
\end{table}

\begin{table}[]
\footnotesize
\centering
\caption{Results with different Reinforcement Learning (RL) methods, where A stands for actor-only version of the algorithm, and AC denotes the complete actor-critic design.}
\label{tab:my-table2}
\resizebox{1.\columnwidth}{!}{
\begin{tabular}{ccccc}
\hline
\multirow{2}{*}{RL Methods} & \multicolumn{2}{c}{Chair-V2} & \multicolumn{2}{c}{Shoe-V2} \\ \cline{2-5} 
 & m@A & m@B & m@A & m@B \\ \hline
Vanilla Policy Gradient & 80.36 & 32.34  & 82.56 & 19.67 \\ 
PPO-AC-Clipping & 81.54  & 33.71 & 83.47 & 20.84 \\ 
PPO-AC-KL Penalty & 80.99  & 32.64 & 83.84 & 20.04 \\ 
PPO-A-KL Penalty & 81.34  & 33.01 & 83.51 & 20.66 \\ 
TRPO &83.21& 33.68 & 83.61 & 20.31   \\ 
PPO-A-Clipping (Ours) & {\bf 85.44} & {\bf 35.09} & {\bf 85.38} & {\bf 21.44}   \\ \hline
\end{tabular}%
}
\end{table}

\begin{table}[]
\footnotesize
\centering
\caption{Results with different candidate reward designs}
 \label{tab:my-table3}
\resizebox{1.\columnwidth}{!}{
\begin{tabular}{ccccc}
\hline
\multirow{2}{*}{Reward Schemes} & \multicolumn{2}{c}{Chair-V2} & \multicolumn{2}{c}{Shoe-V2} \\ \cline{2-5} 
 & m@A & m@B & m@A & m@B \\ \hline
$\mathrm{rank \leq 1\Rightarrow reward = 1}$ & 82.99  & 32.46 & 82.24 & 19.87 \\ 
$\mathrm{rank \leq 5\Rightarrow reward = 1}$ & 81.36  & 31.94 & 81.74 & 19.37 \\
$\mathrm{rank \leq 10\Rightarrow reward = 1}$ & 80.64 & 30.57 &  80.87 &  19.08  \\ 
$\mathrm{-rank}$  &83.71 & 32.84 & 83.81& 20.71 \\
$\mathrm{\frac{1}{\sqrt{rank}}}$  & 83.71 & 33.97 & 83.67 & 20.49 \\ 
$\mathrm{\frac{1}{rank}}$  & 84.33 & 34.11 & 84.07 & 20.54 \\ 
$\mathrm{Ours}$ (Eq.~\ref{reward_3})   & {\bf 85.44} & {\bf 35.09} & {\bf 85.38} & {\bf 21.44}  \\ \hline
\vspace{-.25in}
\end{tabular}%

}
\end{table}

\keypoint{Significance of Global Reward:} While using  our local reward (Eq.~\ref{reward_1}) achieves an excellent rank in early rendering steps, we noticed that the rank of a paired photo might worsen at a certain sketch-rendering step later on, as illustrated in Figure~\ref{fig:Fig3}. As the user attempts to convey more fine-grained detail later on in the process, they may draw some noisy, irrelevant, or outlier stroke that degrades the performance. Our global-reward term in Eq.~\ref{reward_3} alleviates this issue by imposing a monotonically decreasing constraint on the \emph{normalised} Kendall-Tau-distance \cite{knight1966computer} between two successive rank lists over an episode (Figure~\ref{fig:Fig3}). We quantify the identified adverse impacts of inconsistent strokes, via a new metric, termed \textbf{stroke-backlash index}. It is formulated as $\frac{\sum_{t = 2}^{T}\left | min(RP_{t} - RP_{t-1}  , 0)\right |}{T-1}$, where $RP_{t}$ denotes the ranking percentile of paired photo at $t^{th}$ sketch-rendering step which is averaged over all sketch samples in test split. Whenever a newly introduced stroke produces a decline in the ranking percentile, it is considered as a negative performance. Please note that \emph{the lower the value of this index, the better will be the ranking list consistency performance}. We get a decline in stroke-backlash index from $0.0421$ ($0.0451$) to $0.0304$ ($0.0337$) in Chair-V2 (Shoe-V2) dataset when including the global reward. Furthermore, as shown in Table \ref{tab:my-table3}, this global reward term improves the early retrieval performance m@A and (m@B) by 1.11\% (0.98\%) and 1.31\% (0.90\%) for Chair-V2 and Shoe-V2 respectively.
{Instead of imposing the monotonically decreasing constraint over Kendall-Tau-distance that actually considers the relative ranking position of all the photos, we could have imposed the same monotonically decreasing constraint on the specific ranking of the paired-photo only. However, we notice that the stroke-backlash index arises to $0.0416$ ($0.0446$) and overall m@A value decreases by 0.78\% (0.86\%) for Chair-V2 (Shoe-V2), thus justifying the importance of using Kendall-Tau distance in our context.} 

\keypoint{Further Analysis:} (i) We evaluate the performance of our framework with a varying embedding space dimension in Table \ref{tab:my-table4}, confirming our choice of $D=64$. (ii) Instead of using a standalone-trainable diagonal covariance matrix $\Sigma$ for the actor network, we tried employing a separate fully-connected layer to predict the elements of $\Sigma$. However, the m@A and m@B performance deteriorates by 5.64\% (4.67\%) and 4.48\% (3.51\%), for Chair-V2 and (Shoe-V2) datasets respectively. (iii) In the context of on-the-fly FG-SBIR where we possess online sketch-stroke information, a reasonable alternative could be using a recurrent neural network like \cite{collomosse2019livesketch} for modelling the sketch branch instead of CNN. Following SketchRNN's \cite{ha2017neural}  vector representation, a five-element vector is fed at every LSTM unit, and the hidden state vector is passed through a fully connected layer at any arbitrary instant, thus predicting the sketch feature representation. This alleviates the need of feeding a rendered rasterized sketch-image every time. However, replacing the CNN-sketch branch by RNN and keeping rest of the setup unchanged, performance drops significantly.  As a result a top@5 accuracy of 19.62\% (15.34\%) is achieved compared to 76.34\% (65.77\%) in case of CNN for Chair-V2 (Shoe-V2) dataset. (iv) Different people have different stroke orders for sketching. 
Keeping this in mind we conducted an experiment by randomly shuffling stroke orders to check the consistency of our model.
We obtain m@A and m@B values of 85.04\% (34.84\%) and 85.11\% (20.92\%) for Chair-V2 (Shoe-V2) datasets, respectively,  demonstrating our robustness to such variations.

\begin{table}[]
\scriptsize
\centering
\caption{Performance on varying feature-embedding spaces}
\label{tab:my-table4}
\resizebox{1.\columnwidth}{!}{
\begin{tabular}{ c c c c c c c }
\hline
\multirow{2}{*}{} & \multicolumn{3}{c}{Chair-V2} & \multicolumn{3}{c}{Shoe-V2} \\ \cline{2-7} 
                  & m@A      & m@B      & A@5     & m@A      & m@B     & A@5     \\  \hline
32                &   82.61       & 34.67         &       72.67  &     82.94     &     19.61    &      62.31   \\ 
64                & {\bf  85.44}      &   {\bf 35.09 }     &    76.34     &    {\bf 85.38}       &   {\bf 21.44}      &     65.77     \\  
128               &    84.71      &     34.49     &     {\bf 78.61}       &  84.61     &      20.81   &       {\bf 67.64}  \\ 
256               &     81.39     &     31.37     &     77.41      &    80.69    &      19.68   &        66.49 \\ \hline
\vspace{-0.25in}
\end{tabular}%

}
\end{table}

\section{Conclusion}


We have introduced a fine-grained sketch-based image retrieval framework designed to mitigate the practical barriers to FG-SBIR by analysing user sketches on-the-fly, and retrieve photos at the earliest instant. To this end, we have proposed a reinforcement-learning based pipeline with a set of novel rewards carefully designed to encode the `early retrieval' scheme and stabilise the learning procedure against inconsistently drawn strokes. This provides considerable improvement on conventional baselines for on-the-fly FG-SBIR.

{\small
\bibliographystyle{ieee_fullname}
\bibliography{Original_egbib}
}

\end{document}